\documentclass{article}
\usepackage[preprint]{spconf}
\usepackage{amsmath,bm,graphicx,multirow,amssymb}
\usepackage[numbers,sort&compress]{natbib}
\usepackage{booktabs,url}

\allowdisplaybreaks

\title{High Order Recurrent Neural Networks For Acoustic Modelling}
\name{{C. Zhang \& P. C. Woodland}\thanks{Thanks to Mark Gales and the MGB3 team for the MGB3 setup used.}}
\address{Cambridge University Engineering Dept., Trumpington St., Cambridge, CB2 1PZ U.K. \\
  {\small \tt \{cz277,pcw\}@eng.cam.ac.uk}
}
\begin{document}
\ninept
\maketitle
\begin{abstract}
Vanishing long-term gradients are a major issue in training standard recurrent neural networks (RNNs), which can be alleviated by long short-term memory (LSTM) models with memory cells. However, the extra parameters associated with the memory cells mean an LSTM layer has four times as many parameters as an RNN with the same hidden vector size. This paper addresses the vanishing gradient problem using a high order RNN (HORNN) which has additional connections from multiple previous time steps. Speech recognition experiments using British English multi-genre broadcast (MGB3) data showed that the proposed HORNN architectures for rectified linear unit and sigmoid activation functions reduced word error rates (WER) by 4.2\% and 6.3\% over the corresponding RNNs, and gave similar WERs to a (projected) LSTM while using only 20\%--50\% of the recurrent layer parameters and computation.
\end{abstract}
%
%
\toappear{To appear in Proc. ICASSP 2018, April 15-20, 2018, Calgary, Canada}
\copyrightnotice{\copyright IEEE 2018}

\section{Introduction}
\label{sec:intro}
A recurrent neural network (RNN) is an artificial neural network layer where hidden layer outputs from the previous time step form part of the input used to process the current time step \cite{Rumelhart:1986ab,Elman:1990ab}.
This allows information to be preserved through time and is well suited to sequence processing problems, such as acoustic and language modelling for automatic speech recognition \cite{Robinson:1996ab,Mikolov:2012ab}. However, training RNNs with sigmoid activation functions by gradient descent can be difficult. The key issues are \textit{exploding and vanishing gradients} \cite{Bengio:1994ab},
{\em i.e.}, the long-term gradients, which are back-propagated through time, can either continually increase (explode) or decrease to zero. This causes RNN training to either fail to capture long-term temporal relations or for standard update steps to put parameters out of range.

Many methods have been proposed to solve the gradient exploding and vanishing problems. While simple gradient clipping has been found to work well in practice to prevent gradient exploding \cite{Mikolov:2012ab}, 
circumventing vanishing gradients normally requires more sophisticated strategies \cite{Pascanu:2013ab}. For instance \cite{Sutskever:2011ab} uses Hessian-Free training which makes use of second-order derivative information. 
Modifying the recurrent layer structure is another approach. The use of both rectified linear unit (ReLU) and sigmoid activation functions with trainable amplitudes were proposed to maintain the magnitude of RNN long-term gradients \cite{Salinas:1996ab,Hahnloser:1998ab,Goh:2003ab}. A gating technique is used  in the long short-term memory (LSTM) model where additional parameters implement a memory circuit which can remember long-term information from the recurrent layer \cite{Hochreiter:1997ab}. 
A model similar to the LSTM is the gated recurrent unit \cite{Chung:2014ab}. More recently, additional residual \cite{He:2016ab} and highway connections \cite{Srivastava:2015ab} were proposed to train very deep feed-forward models, 
which allows gradients to pass more  easily through many layers. 
Various similar ideas have been applied to recurrent models 
\cite{Srivastava:2016ab,Zhang:2016ab,Tian:2016ab,vandenOord:2016ab,Pundak:2017ab,Kim:2017ab}. Among these
 approaches, the LSTM has recently  become the dominant type of recurrent architecture. However LSTMs, due to the extra parameters associated with gating,  use four times more parameters
as standard RNNs with the same hidden layer size, which significantly increases storage and computation in both training and testing. 

In this paper, we propose another  RNN modification, the high order RNN (HORNN), as an alternative to the LSTM. It handles vanishing gradients by adding  connections from hidden state values at multiple previous time steps to the  RNN input. By interpreting the RNN layer hidden vector as a continuous valued hidden state, the connections are termed high order since they introduce dependencies on  multiple previous hidden states.
Acoustic modelling using HORNNs is investigated for both sigmoid and ReLU activation functions. In the sigmoid case, it is found that additional high order connections are beneficial.
Furthermore, analogous to the projected LSTM (LSTMP) \cite{Sak:2014ab}, a linear recurrent projection layer can be used by HORNNs to reduce the number of parameters, which results in the projected HORNN (HORNNP). 
Experimental results show that  the HORNN/HORNNP (both sigmoid and ReLU) have similar word error rates (WERs) to LSTM/LSTMP models with the same hidden vector size, while using fewer than half the parameters and computation.
Furthermore, HORNNs were also found to outperform RNNs with residual connections in terms of both speed and WER.

This paper is organised as follows. Section~\ref{sec:basic} reviews RNN and LSTM models. The (conditional) Markov property of RNNs is described in Sec.~\ref{sec:highorder}, which leads to HORNNs and architectures for both sigmoid and ReLU activation functions. The experimental setup and results are given in Sec.~\ref{sec:expsetup} and Sec.~\ref{sec:expresults}, followed by conclusions.

\section{RNN and LSTM models}
\label{sec:basic}

In this paper, an RNN refers to an \textit{Elman network} \cite{Elman:1990ab}  that produces its output hidden vector at step $t$, $\mathbf{h}_t$, based on the previous output $\mathbf{h}_{t-1}$ and the current input $\mathbf{x}_t$ by
\begin{align}
	\label{eq:rnn1}
	\mathbf{h}_t=f(\mathbf{a}_t)=f(\mathbf{W}\mathbf{x}_t+\mathbf{U}\mathbf{h}_{t-1}+\mathbf{b}),
\end{align} 
where $\mathbf{W}$ and $\mathbf{U}$ are the weights, $\mathbf{b}$ is the bias, and $f(\cdot)$ and $\mathbf{a}_t$ are the activation function and its input activation value. In general, $\mathbf{h}_t$ is processed by a number of further  layers to obtain the final network output.
It is well known that when $f(\cdot)$ is the sigmoid denoted $\sigma(\cdot)$, RNNs suffer from the  vanishing gradient issue since 
\begin{align*}
\dfrac{\partial{\sigma}(\mathbf{a}_t)}{\partial\mathbf{a}_t}={\sigma}\left(\mathbf{a}_t)(1-{\sigma}(\mathbf{a}_t)\right)\leqslant\dfrac{1}{4},
\end{align*} 
which enforces gradient magnitute reductions in backpropagation \cite{Robinson:1996ab}. Note that ReLU RNNs suffer less from this issue. 

In contrast to a standard RNN, the LSTM model resolves gradient vanishing by using an additional linear state $\mathbf{c}_t$ at each step of the  sequence, which can be viewed as a memory cell. 
At each step, a new cell candidate $\tilde{\mathbf{c}}_t$ is created to encode the information from the current step. $\mathbf{c}_t$ is first updated by interpolating $\mathbf{c}_{t-1}$ with $\tilde{\mathbf{c}}_t$ based on the forget gate $\mathbf{f}_t$ and input gate $\mathbf{i}_t$, and then converted to the LSTM hidden state by transforming with hyperbolic tangent (tanh) and scaling by the output gate $\mathbf{o}_t$. This procedure simulates a memory circuit where $\mathbf{f}_t$, $\mathbf{i}_t$, and $\mathbf{o}_t$ are analogous to its logic gates \cite{Hochreiter:1997ab}. More specifically, an LSTM layer step $t$ is evaluated as
\begin{align*}
	\mathbf{i}_t&= {\sigma}(\mathbf{W}_i\mathbf{x}_t+\mathbf{U}_i\mathbf{h}_{t-1}+\mathbf{V}_i\odot\mathbf{c}_{t-1}+\mathbf{b}_i),\\
	\mathbf{f}_t&= {\sigma}(\mathbf{W}_f\mathbf{x}_t+\mathbf{U}_f\mathbf{h}_{t-1}+\mathbf{V}_f\odot\mathbf{c}_{t-1}+\mathbf{b}_f),\\
	\tilde{\mathbf{c}}_t&= {\tanh}(\mathbf{W}_c\mathbf{x}_t+\mathbf{U}_c\mathbf{h}_{t-1}+\mathbf{b}_c),{~}\mathbf{c}_t=\mathbf{f}_t\odot\mathbf{c}_{t-1}+\mathbf{i}_t\odot\tilde{\mathbf{c}}_t,\\
	\mathbf{o}_t&= {\sigma}(\mathbf{W}_o\mathbf{x}_t+\mathbf{U}_o\mathbf{h}_{t-1}+\mathbf{V}_o\odot\mathbf{c}_{t}+\mathbf{b}_o),\\
	\mathbf{h}_t&=\mathbf{o}_t\odot{\tanh}(\mathbf{c}_t),
\end{align*} 
where $\odot$ represents element-wise product, and the $\mathbf{V}$ matrices are diagonal which serve as a ``peephole''. Although LSTMs work very well on a large variety of tasks, it is computationally very expensive. 
The representations for each temporal step, $\tilde{\mathbf{c}}_t$, are extracted in the same way as the RNN $\mathbf{h}_t$. However, the additional cost  of finding $\mathbf{c}_t$ over $\tilde{\mathbf{c}}_t$, requires three times the computation and parameter storage since  $\mathbf{i}_t$, $\mathbf{f}_t$, and $\mathbf{o}_t$ all need to be calculated.

\section{High Order RNN Acoustic Models}
\label{sec:highorder}
In this section, HORNNs are proposed by relaxing the first-order Markov conditional independence constraint.

\subsection{Markov Conditional Independence}
\label{ssec:1stordermarkov}
The posterior probability of the $T$ frame label sequence $y_{1:T}$ given the $T$ frame input sequence $\mathbf{x}_{1:T}$ can be found by integrating over all possible continuous hidden state sequences $\tilde{\mathbf{h}}_{1:T}$ 
\begin{align*}
	&P(y_{1:T}|\mathbf{x}_{1:T})=\int P(y_{1:T}|\tilde{\mathbf{h}}_{1:T},\mathbf{x}_{1:T})p(\tilde{\mathbf{h}}_{1:T}|\mathbf{x}_{1:T}) \,d \tilde{\mathbf{h}}_{1:T}\\
	&=\int\prod^{T}_{t=1}P(y_t|y_{1:t-1},\tilde{\mathbf{h}}_{1:T},\mathbf{x}_{1:T})p(\tilde{\mathbf{h}}_t|\tilde{\mathbf{h}}_{1:t-1},\mathbf{x}_{1:T})\,d \tilde{\mathbf{h}}_{1:T}.
\end{align*} 
When implemented using an RNN,
\begin{align*}
P(y_t|y_{1:t-1},\tilde{\mathbf{h}}_{1:T},\mathbf{x}_{1:T})=P(y_t|\tilde{\mathbf{h}}_{t}),
\end{align*}
which is produced by the layers after the RNN layer. From Eqn.~\eqref{eq:rnn1}, $\tilde{\mathbf{h}}_t$ depends only on $\tilde{\mathbf{h}}_{t-1}$ and $\mathbf{x}_t$, {\em i.e.},
\begin{align}
	\label{eq:rnn2}
	p(\tilde{\mathbf{h}}_t|\tilde{\mathbf{h}}_{1:t-1},\mathbf{x}_{1:T})=p(\tilde{\mathbf{h}}_{t}|\tilde{\mathbf{h}}_{t-1},\mathbf{x}_t).
\end{align}	
Since the initial hidden state is given (often set to $\mathbf{h}_0=\mathbf{0}$), all subsequent states $\mathbf{h}_{1:T}$ are determined by Eqn.~\eqref{eq:rnn1}, 
which means  $p(\tilde{\mathbf{h}}_{t}|\mathbf{h}_{t-1},\mathbf{x}_t)$ is a Kronecker delta function
\begin{align*}
    p(\tilde{\mathbf{h}}_t|\mathbf{h}_{t-1},\mathbf{x}_{t})=\left\{\begin{array}{ll}
      1 & \text{if}{~~}\tilde{\mathbf{h}}_t=\mathbf{h}_{t}\\
   	  0 & \text{otherwise}
   	  \end{array}\right..
\end{align*}
Hence $P(y_{1:T}|\mathbf{x}_{1:T})=\prod^{T}_{t=1}P(y_t|\mathbf{h}_t=f(\mathbf{W}\mathbf{x}_t+\mathbf{U}\mathbf{h}_{t-1}+\mathbf{b}))$.

Eqn.~\eqref{eq:rnn2} is the 1st-order \textit{Markov conditional independence property} 
\cite{Bengio:1993ab}. It means that the current state $\mathbf{h}_t$ depends only on its immediately  preceding state $\mathbf{h}_{t-1}$ and the current input $\mathbf{x}_{t}$. This property differs from the  1st-order \textit{Markov property} by also conditioning on $\mathbf{x}_{t}$.\footnote{For language modelling, $\mathbf{h}_t$ has the standard Markov property as the RNN models $P(y_{1:T})$ without conditioning on $\mathbf{x}_{1:T}$.} Note that this property also applies to bidirectional RNNs  \cite{Schuster:1997ab}, which is easy to show by defining a new hidden state $\mathbf{h}^{\text{bid}}_t=\{\mathbf{h}^{\text{fwd}}_t,\mathbf{h}^{\text{bwd}}_t\}$, where $\mathbf{h}^{\text{fwd}}_t$ and $\mathbf{h}^{\text{bwd}}_t$ are the forward and backward RNN hidden states.

\subsection{HORNNs for Sigmoid and ReLU Activation Functions}
\label{ssec:highorder}
In this paper, the gradient vanishing issue is tackled by relaxing the first-order Markov conditional independence constraint. Hence, not only the direct preceding state $\mathbf{h}_{t-1}$ but also previous states $\mathbf{h}_{t-n} (n>1)$  are used when calculating $\mathbf{h}_t$. This adds additional high order connections to the RNN architecture and results in a HORNN. From a training perspective, including high order states creates shortcuts for backpropagation to allow additional long-term information to flow more easily. 
Specifically, the gradients w.r.t. $\mathbf{h}_{t-1}$ of a general $n$-order RNN can be obtained by
\begin{align}
\label{eq:hornn1}
\dfrac{\partial\mathcal{F}}{\partial\mathbf{h}_{t-1}}=\sum_{i=1}^{n}\dfrac{\partial\mathcal{F}}{\partial\mathbf{h}_{t-i-1}}\dfrac{\partial\mathbf{h}_{t-i-1}}{\partial\mathbf{h}_{t-1}},
\end{align}
where $\mathcal{F}$ is the training criterion. 
For $n>1$, Eqn.~\eqref{eq:hornn1} sums multiple terms to prevent the gradient vanishing. 
From an inference (testing) perspective, an RNN assumes sufficient past temporal information has been  embedded in the representation $\mathbf{h}_{t-1}$, but using a fixed sized $\mathbf{h}_{t-1}$, means that information from 
distant long-term steps may not be properly integrated with new short-term information. The HORNN architecture allows more direct access to the past long-term information. 

There are many alternative ways of using $\mathbf{h}_{t-n}$ in the calculation of $\mathbf{h}_{t}$  in the HORNN framework. This paper assumes that the high order connections are linked to the input at step $t$. 
It was found to be sufficient to use only one high order connection at the input, {\em i.e.}
\begin{align}
	\label{eq:hornn2}
	\mathbf{h}_t=f(\mathbf{W}\mathbf{x}_t+\mathbf{U}_{1}\mathbf{h}_{t-1}+\mathbf{U}_n\mathbf{h}_{t-n}+\mathbf{b}).
\end{align}
Here $\mathbf{h}_{t-n}$ can be viewed as a kind of ``memory'' whose temporal resolution is modified by $\mathbf{U}_n$.
From our experiments the structure in Eqn.~\eqref{eq:hornn2} allowed ReLU HORNNs to give similar WERs to LSTMs. 
However, when using sigmoid HORNNs, a slightly different structure is needed to reach a similar WER. This has an extra high order connection from $\mathbf{h}_{t-m}$ to the sigmoid function input, {\em i.e.}
\begin{align}
	\label{eq:hornn3}
	\mathbf{h}_t=f(\mathbf{W}\mathbf{x}_t+\mathbf{U}_1\mathbf{h}_{t-1}+\mathbf{U}_n\mathbf{h}_{t-n}+\mathbf{h}_{t-m}+\mathbf{b}).
\end{align} 
Here, $\mathbf{h}_{t-m}$ is directly added to the sigmoid input without impacting the temporal resolution at $t$ since $\mathbf{h}_{t-m}$ is from a previous sigmoid output. 
Eqns.~\eqref{eq:hornn2} and \eqref{eq:hornn3} are used for ReLU and sigmoid HORNNs throughout the paper.

\subsection{Parameter Control using Matrix Factorisation}
\label{ssec:factorise}
Comparing Eqns.~\eqref{eq:hornn2} and \eqref{eq:hornn3} to Eqn.~\eqref{eq:rnn1}, HORNN increases the number of RNN layer parameters from $(D_x+D_h)D_h+D_h$ to $(D_x+2D_h)D_h+D_h$, where $D_x$ and $D_h$ are the sizes of $\mathbf{x}_t$ and $\mathbf{h}_t$. One method to reduce the  increase in  parameters is to project the hidden state vectors to some a dimension  $D_p$ with  a recurrent linear projection $\mathbf{P}$ \cite{Sak:2014ab}. This factorises $\mathbf{U}_1$ and $\mathbf{U}_n$ in Eqns.~\eqref{eq:hornn2} and \eqref{eq:hornn3} to $\mathbf{U}_{p1}\mathbf{P}$ and $\mathbf{U}_{pn}\mathbf{P}$ with a \textit{low-rank approximation}. The projected HORNNs (denoted by HORNNP) for ReLU and sigmoid activations are hence defined as
\begin{align}
	\label{eq:hornnp1}
	\mathbf{h}_t&=f(\mathbf{W}\mathbf{x}_t+\mathbf{U}_{p1}\mathbf{P}\mathbf{h}_{t-1}+\mathbf{U}_{pn}\mathbf{P}\mathbf{h}_{t-n}+\mathbf{b})
	\end{align}
	and
	\begin{align}
	\label{eq:hornnp2}
	\mathbf{h}_t&=f(\mathbf{W}\mathbf{x}_t+\mathbf{U}_{p1}\mathbf{P}\mathbf{h}_{t-1}+\mathbf{U}_{pn}\mathbf{P}\mathbf{h}_{t-n}+\mathbf{h}_{t-m}+\mathbf{b}),
\end{align}
and the number of parameters used is $D_hD_p+(D_x+2D_p)D_h+D_h$. The resulting parameter reduction ratio is approximately $2D_h/3D_p$ (given $D_h>D_p\gg D_x$). Note that the same idea was used by the projected LSTM (LSTMP) to factorise $\mathbf{U}_i$, $\mathbf{U}_f$, $\mathbf{U}_c$, and $\mathbf{U}_o$ \cite{Sak:2014ab}, which reduces the  number of LSTM parameters from $4(D_x+D_h)D_h+7D_h$ to $D_hD_p+4(D_x+D_p)D_h+7D_h$.

Next we compare the computational complexity of LSTMs and HORNNs. Given that  multiplying a $l\times m$ matrix by a $m\times n$ matrix ($l\neq m\neq n$) requires  $lmn$ multiply-adds, and ignoring all element-wise operations, the testing complexity for a HORNNP layer is $\mathcal{O}(T(D_x+3D_p)D_h)$, whereas  for an LSTMP it is $\mathcal{O}(TD_hD_p+4T(D_x+D_p)D_h)$. This shows that HORNNPs use less than 3/5 of the calculations of LSTMPs.  It has been found that HORNNPs often result in a 50\% speed up over LSTMPs in our current HTK implementation \cite{Zhang:2015ef,Young:2015ab,Zhang:2017ab}.

\subsection{Related Work}
\label{ssec:relatedwork}
After independently developing the HORNN  for acoustic modelling, we found that similar ideas had previously been applied to rather different tasks \cite{Lin:1996ab,Tino:2004ab,Sutskever:2010ab,Soltani:2016ab,Huang:2017ab}. However, both the research focus and model architectures were different to this paper. In particular, the model proposed in \cite{Lin:1996ab,Soltani:2016ab} is equivalent to Eqn.~\eqref{eq:hornn2} without subsampling the high order hidden vectors, and \cite{Huang:2017ab} applied that model to TIMIT phone recognition. Furthermore, previous studies didn't  discuss the high order connections  in the Markov property framework.

Adding $\mathbf{h}_{t-m}$ to the input of the sigmoid function in Eqn.~\eqref{eq:hornn3} is similar to the residual connection in residual networks \cite{He:2016ab}. A residual RNN (ResRNN) with a \textit{recurrent kernel depth} of two ($d=2$) can be written as
\begin{align}
	\label{eq:resrnn1}
	\mathbf{h}_t=f(\mathbf{U}_{d2}f(\mathbf{W}\mathbf{x}_t+\mathbf{U}_{d1}\mathbf{h}_{t-1}+\mathbf{b})+\mathbf{h}_{t-m}),
\end{align} 
where $m=1$ \cite{Tian:2016ab}.
Another related model is the recent residual memory network \cite{Baskar:2017ab}, which can be viewed as an unfolded HORNN defined in Eqn.~\eqref{eq:hornn2} with $\mathbf{U}_{1}$ and $\mathbf{b}$ being zero, $\mathbf{W}$ being distinct untied parameters in each unfolded layer, and $n\geqslant1$ being any positive integer. 
In addition, since highway networks \cite{Srivastava:2015ab} can be viewed as a generalised form of the residual networks, highway RNNs and LSTMs are also related to this work \cite{Srivastava:2016ab,Pundak:2017ab}. 
Note that it is also possible to combine the residual and highway ideas with HORNNs by increasing the recurrent depth.

\section{Experimental Setup}
\label{sec:expsetup}

The proposed HORNN models were evaluated by training systems on multi-genre broadcast (MGB) data from the MGB3 speech recognition challenge task \cite{mgb3website,Bell:2015ab}. The audio is from BBC TV programmes covering a range of genres.
A 275 hour (275h)  full training set was selected from 750 episodes 
where the sub-titles have a phone matched error rate $<40\%$ compared to the lightly supervised output \cite{Lanchantin:2016ab} which was used as training supervision. 
A 55 hour (55h) subset was sampled at the utterance level from the 275h set. A 63k word vocabulary \cite{Richmond:2010ab} was used with a trigram word level language model (LM) estimated from both the acoustic transcripts and a separate 640 million word MGB subtitle archive.
The test set, $\textbf{dev17b}$, contains 5.55 hours of audio data and 5,201 manually segmented utterances from 14 episodes of 13 shows. This is a subset of the official full development set (\textbf{dev17a}) with data that overlaps training and test sets excluded. System outputs were evaluated  with confusion network decoding (CN)  \cite{Evermann:2000ab} as well as 1-best Viterbi decoding.

All experiments were conducted with an extended version of HTK 3.5 \cite{Young:2015ab,Zhang:2015ef}. The LSTM was implemented following \cite{Sak:2014ab}. A 40d log-Mel filter bank analysis was used and expanded to an 80d vector with its $\Delta$ coefficients. The data was normalised at the utterance level for mean and at the show-segment level for variance \cite{Woodland:2015ab}. 
The inputs at each recurrent model time step were single frames delayed for 5 steps \cite{Sak:2014ab,Li:2017ab}.
All models were trained using the cross-entropy criterion and  frame-level shuffling used. All recurrent models were unfolded for 20 time steps, and the gradients of the shared parameters were normalised by dividing by the sharing counts \cite{Zhang:2015ef}. 
The maximum parameter changes were constrained by update value clipping with a threshold of 0.32 for a minibatch with 800 samples.

 About 6k/9k decision tree clustered triphone tied-states along with GMM-HMM/DNN-HMM system training alignments were used for the 55h/275h training sets. 
One hidden layer with the same dimension as $\mathbf{h}_t$ was added between the recurrent and output layers to all models.
The NewBob$^{+}$ learning rate scheduler \cite{Zhang:2015ef,Zhang:2017ab} was used to train all models with the setup from our previous MGB systems \cite{Woodland:2015ab}.
An initial learning rate of $5\times10^{-4}$ was used for all ReLU models, while an initial rate of $2\times10^{-3}$ was used to train all the other models. Since regularisation plays an important role in RNN/LSTM training,  weight decay factors were carefully tuned to maximise the performance of each system. 

\section{Experimental Results}
\label{sec:expresults}

\subsection{55 Hour Single Layer HORNN Experiments}
\label{sssec:expsub1}

\begin{figure}[htb]
\begin{minipage}[b]{1.0\linewidth}
  \centering
  \centerline{\includegraphics[width=8.5cm]{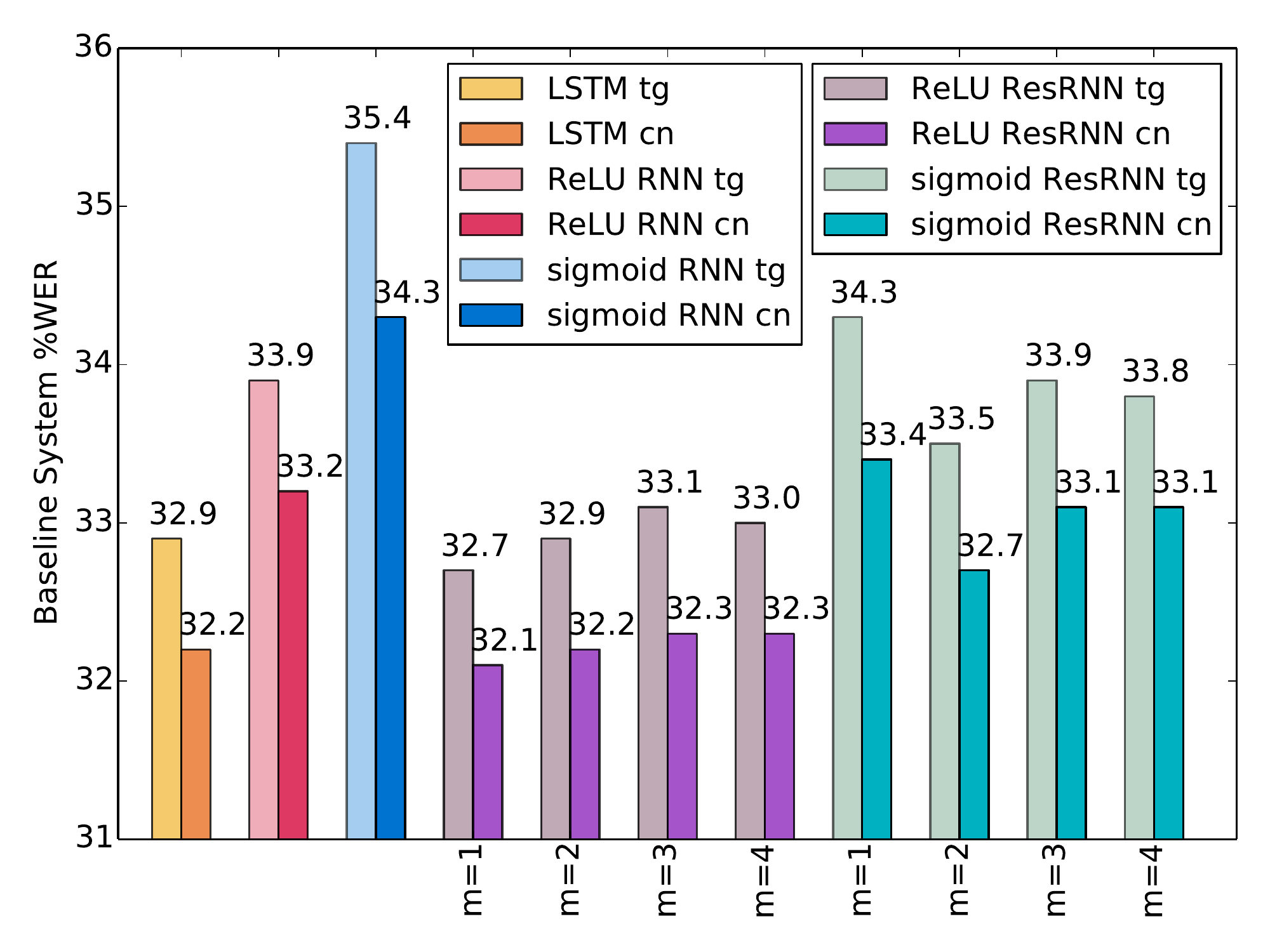}}
\end{minipage}
%
%
\begin{minipage}[b]{1.0\linewidth}
  \centering
  \vspace{-2mm}
  \centerline{\includegraphics[width=8.5cm]{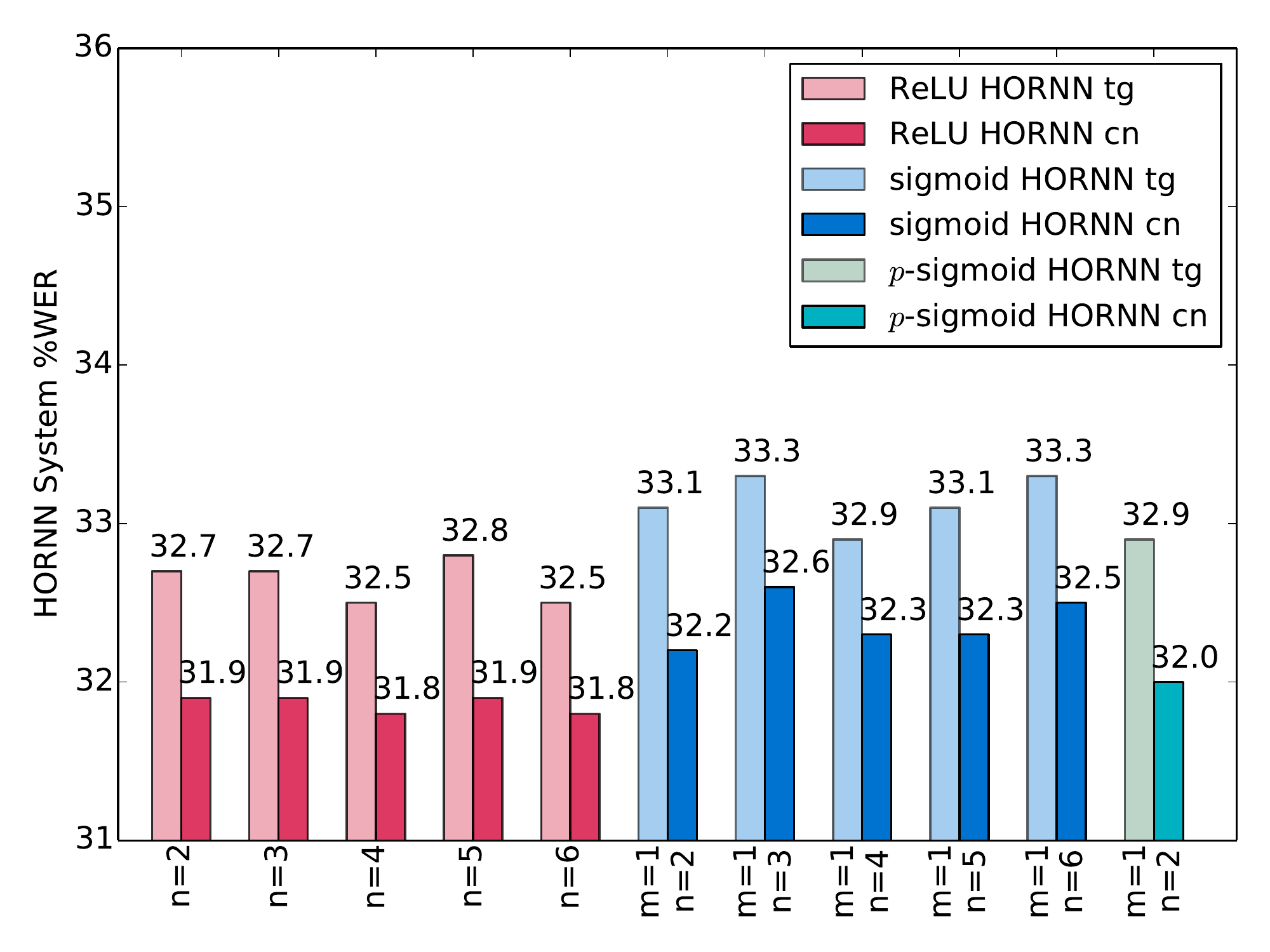}}
\end{minipage}
\vspace{-8mm}
\caption{\it \%WERs of 55h systems  on dev17b. Systems use a trigram LM with Viterbi decoding (tg) or CN decoding (cn). }
\label{fig:res1}
\vspace{-4mm}
\end{figure}

Initial experiments studied various HORNN architectures in order to investigate suitable values of $n$ for the ReLU model in Eqn.~\eqref{eq:hornn2}, and for both $m$ and $n$ for the sigmoid model in Eqn.~\eqref{eq:hornn3}. To save computation,  the 55h subset was used for training. All models had one recurrent layer with the $\mathbf{h}_t$ size fixed to 500. An LSTM and a standard RNN were created as baselines, which had 1.16M and 0.29M parameters in the recurrent layers respectively. A ResRNN, defined by Eqn.~\eqref{eq:resrnn1} was also tested as an additional baseline  using both ReLU and sigmoid functions.\footnote{This is also the first time to apply such ResRNNs to acoustic modelling.} 
ResRNNs had the same number of parameters (0.54M) as the HORNNs. Note that rather than the standard case with $m=1$ \cite{Tian:2016ab}, $m\in[1,4]$ were examined which falls into the high order framework when $m>1$. For HORNNs, $n\in[2,6]$ were tested; $m$ was fixed to 2 for all sigmoid HORNNs. From the results shown in Figure~\ref{fig:res1}, the LSTM gives lower WERs  than a standard RNN, but the ReLU ResRNN with $m$ set to 1 or 2 had a similar WER to the LSTM. 

ReLU HORNNs gave  WERs at least as low as the LSTM and the best ReLU ResRNN systems. Sigmoid HORNNs gave better WERs than sigmoid ResRNNs and similar WERs to those from the LSTM. 
The performance can be further improved by using $p$-sigmoid \cite{Zhang:2015cd} as the HORNN activation function which associates a linear scaling factor to each recurrent layer output unit and makes it more similar to a ReLU. 
In addition, HORNNs were faster than both LSTM and ResRNNs. ResRNNs were slightly slower than HORNNs since the second matrix multiplication depends on the first one at each recurrent step. 
For the rest of the experiments, all ReLU HORNNs used $n=4$ , and all sigmoid HORNNs used $m=1$ and $n=2$.

      \begin{table}[h]
        \vspace{2mm}
        \centerline{
          \begin{tabular}{clccccc}
            \toprule
			ID & System & $D_h$ & $D_p$ & tg & cn \\
            \midrule
			L$^{\text{55h}}_1$ & 1L LSTMP & 500 & 250 & 32.9 & 32.1 \\	
			L$^{\text{55h}}_2$ & 1L LSTMP & 600 & 300 & 32.7 & 32.0 \\	
			L$^{\text{55h}}_3$ & 2L LSTMP & 500 & 250 & 31.3 & 30.6 \\	
			\cmidrule{1-6}
			S$^{\text{55h}}_{1}$ & 1L sigmoid HORNNP & 500 & 250 & 32.8 & 31.9 \\	
						S$^{\text{55h}}_{2}$ & 1L sigmoid HORNNP & 500 & 125 & 33.0 & 32.1 \\	
			S$^{\text{55h}}_{3}$ & 1L sigmoid HORNNP & 800 & 400 & 31.6 & 30.9 \\	
						S$^{\text{55h}}_{4}$ & 2L sigmoid HORNNP & 500 & 250 & 31.4 & 30.7 \\	
			\cmidrule{1-6}
						R$^{\text{55h}}_{1}$ & 1L ReLU HORNNP & 500 & 250 & 32.0 & 31.4 \\	
						R$^{\text{55h}}_{2}$ & 1L ReLU HORNNP & 500 & 125 & 32.5 & 31.8 \\	
						R$^{\text{55h}}_{3}$ & 1L ReLU HORNNP & 800 & 400 & 31.4 & 30.7 \\	
			R$^{\text{55h}}_{4}$ & 2L ReLU HORNNP & 500 & 250 & 31.4 & 30.7 \\	
            \bottomrule
          \end{tabular}
        }
		\caption{\label{tab:55h} {\it \%WERs for various 55h system on dev17b.  Systems use a trigram LM with Viterbi decoding (tg) or CN decoding (cn).}}
		\vspace{-5mm}
      \end{table}

\subsection{Projected and Multi-Layered HORNN Results}
\label{sssec:expsub2}
Next, projected LSTMs and projected HORNNs were compared. 
First, $D_h$ (the size of $\mathbf{h}_t$) and $D_p$ (the projected vector size) were fixed to 500 and 250 respectively for the single recurrent layer (1L) LSTMP and HORNNP models. 
The LSTMP baseline L$^{\text{55h}}_1$ had 0.79M parameters and HORNNP system S$^{\text{55h}}_{1}$ and R$^{\text{55h}}_{1}$ had 0.42M parameters. From Table~\ref{tab:55h}, the HORNNPs have similar WERs to the LSTMP. By further reducing $D_p$ to 250, the HORNN systems, S$^{\text{55h}}_{2}$ and R$^{\text{55h}}_{2}$, reduced  the number of  parameters to 0.23M and gave similar WERs to LSTM and LSTMP (L$^{\text{55h}}_1$) with only 20\% and 29\% of the recurrent layer parameters.

The values of  $D_h$ and $D_p$ for HORNNs were increased to 800 and 400 respectively to make the overall number of recurrent layer parameters (1.02M) closer to that of the 500d LSTM (1.16M). This produced system S$^{\text{55h}}_{3}$ and R$^{\text{55h}}_{3}$. The LSTMP was also modified to $D_h=600$ and $D_p=300$ to have 1.10M parameters. From the results in Table~\ref{tab:55h}, S$^{\text{55h}}_{3}$ and R$^{\text{55h}}_{3}$ both outperformed L$^{\text{55h}}_2$ by a margin since the 800d representations embed more accurate temporal information than with 600d. The $p$-sigmoid function was not used for HORNNPs since the linear projection layer also scales $\mathbf{h}_t$. 

Finally, the  LSTMP and HORNNP were compared by stacking  another recurrent layer. With two recurrent layers (2L) of $D_h=500$ and $D_p=250$, the 2L HORNNP systems S$^{\text{55h}}_{4}$ and R$^{\text{55h}}_{4}$ had 0.92M parameters and still produced similar WERs to the 2L LSTMP system L$^{\text{55h}}_3$ (with 1.91M parameters).
These results indicate that rather than spending most of the calculations on maintaining the LSTM memory cell, it is more effective to use HORNNs and use the computational budget for extracting better temporal representations using wider and deeper recurrent layers.

\subsection{Experiments on 275 Hour Data Set}
\label{sssec:expsub3}
To ensure that the previous results scale to a significantly larger training set, some selected LSTMP and HORNNP systems were built on the full 275h set. Here $D_h$ and $D_p$   were set to 1000 and 500, which increased the number of recurrent layer parameters to better model the full training set. 
From Table~\ref{tab:275h}, for both single recurrent layer and two recurrent layer architectures, HORNNs still produced similar  WERs to the corresponding LSTMPs. This validates our previous finding on a larger data set that the proposed HORNN structures can work as well as the widely used LSTMs on acoustic modelling by using far fewer parameters. In addition, along with the multi-layered structure, HORNNs can also be applied to other kinds of recurrent models by replacing RNNs and LSTMs, such as the bidirectional \cite{Schuster:1997ab} and grid \cite{Li:2017ab,Kalchbrenner:2016ab,Kreyssig:2018ab} structures \textit{etc}. 
Finally, a 7 layer (7L) sigmoid DNN system, D$^{\text{275h}}_1$, was built following \cite{Woodland:2015ab} as a reference.

%
%

      \begin{table}[th]
        \vspace{2mm}
        \centerline{
          \begin{tabular}{clccccc}
            \toprule
			ID & System & $D_h$ & $D_p$ & tg & cn \\
            \midrule
						L$^{\text{275h}}_1$ & 1L LSTMP & 1000 & 500 & 26.5 &  26.0 \\	
						S$^{\text{275h}}_1$ & 1L sigmoid HORNNP & 1000 & 500 & 26.4 & 25.8 \\	
						R$^{\text{275h}}_1$ & 1L ReLU HORNNP & 1000 & 500 & 26.4 & 25.9 \\	
			\cmidrule{1-6}
						L$^{\text{275h}}_3$ & 2L LSTMP & 1000 & 500 & 25.7 &  25.2 \\	
						S$^{\text{275h}}_4$ & 2L sigmoid HORNNP & 1000 & 500 & 25.6 &  25.2 \\	
						R$^{\text{275h}}_4$ & 2L ReLU HORNNP & 1000 & 500 & 25.3 & 25.0 \\	
			\cmidrule{1-6}
			D$^{\text{275h}}_1$ & 7L sigmoid DNN & 1000 &  & 28.4 & 27.5 \\ 
            \bottomrule
          \end{tabular}
        }
         	\caption{\label{tab:275h}{\it \%WERs for a selection of 275h system on dev17b.  Systems use a trigram LM with Viterbi decoding (tg) or CN decoding (cn).}}
		\vspace{-5mm}
      \end{table}


\section{CONCLUSIONS}
\label{sec:conclusion}
This paper proposed the use of HORNNs for acoustic modelling to address the vanishing gradient problem in training recurrent neural networks. 
 Two different architectures were proposed to cover both ReLU and sigmoid activation functions. These yielded 4\%-6\%
WER reductions over the standard RNNs with the same activation function. 
Furthermore, additional structures were investigated: reducing the number of HORNN parameters with a linear recurrent projected layer; and adding another recurrent layer. In all cases, compared to the projected LSTMs and the residual RNNs, 
it was shown that HORNNs gave similar WER performance while being significantly more efficient in computation and storage. 
When the savings in parameter number and computation are used to implement wider or deeper recurrent layers, (projected) HORNNs gave a 4\% relative reduction in WER over the comparable (projected) LSTMs .

\vfill\pagebreak

\renewcommand{\bibsection}{}
\section{REFERENCES}
\eightpt


\begin{thebibliography}{10}
	
\providecommand{\newblock}{\relax}

\bibitem{Rumelhart:1986ab}
D.E. Rumelhart, J.L. McClelland, \& the PDP Research Group
\newblock {\em Parallel Distributed Processing: Explorations in the Microstructure of Cognition, Volume 1: Foundations},
\newblock {MIT Press},
1986.


\bibitem{Elman:1990ab}
 J.L.~Elman,
 \newblock ``Finding structure in time",
 \newblock {\em Cognitive Science},
   vol. 14, pp. 179--211, 1990.

\bibitem{Robinson:1996ab}
T. Robinson, M. Hochberg and S. Renals. 
\newblock {``The use of recurrent neural networks in continuous speech recognition"},
\newblock In {\em Automatic Speech and Speaker Recognition}, pp. 233--258, Springer, 1996.

\bibitem{Mikolov:2012ab}
 T.~Mikolov,
\newblock {\em Statistical Language Models based on Neural Networks},
\newblock  {Ph.D. thesis}, Brno University of Technology,  Brno, Czech Republic, 2012.



\bibitem{Bengio:1994ab}
 Y.~Bengio, P.~Simard, \& P.~Frasconi,
 \newblock ``Learning long-term dependencies with gradient descent is difficult",
 \newblock {\em IEEE Transactions on Neural Networks},
   vol. 5, pp. 157--166, 1994.


\bibitem{Pascanu:2013ab}
R.~Pascanu, T.~Mikolov, \& Y.~Bengio,
\newblock ``On the difficulty of training recurrent neural networks",
\newblock  {\em Proc. ICML}, Atlanta, 2013.

\bibitem{Sutskever:2011ab}
I.~Sutskever, J.~Martens, \& G.~Hinton,
\newblock ``Generating text with recurrent neural networks",
\newblock  {\em Proc. ICML}, New York, 2011.

\bibitem{Salinas:1996ab}
E.~Salinas \& L.F.~Abbott,
 \newblock ``A model of multiplicative neural responses in parietal cortex",
 \newblock {\em Proc. National Academy of Science U.S.A.},
   vol. 93, pp. 11956--11961, 1996.

\bibitem{Hahnloser:1998ab}
R.L.T.~Hahnloser,
 \newblock ``On the piecewise analysis of networks of linear threshold neurons",
 \newblock {\em Neural Networks},
   vol. 11, pp. 691--697, 1998.
   
\bibitem{Goh:2003ab}
S.L.~Goh \& D.P.~Mandic
 \newblock ``Recurrent neural networks with trainable amplitude of activation functions",
 \newblock {\em Neural Networks},
   vol. 16, pp. 1095--1100, 2003.

\bibitem{Hochreiter:1997ab}
S.~Hochreiter \& J.~Schmidhuber,
 \newblock ``Long short-term memory",
 \newblock {\em Neural Computation},
   vol. 9, pp. 1735--1780, 1997.
   


\bibitem{Chung:2014ab}
J.~Chung, C.~Gulcehre, K.H.~Cho, \& Y.~Bengio,
\newblock ``Empirical evaluation of gated recurrent neural networks on sequence modeling",
\newblock  {\em arXiv.org}, 1412.3555, 2014.

\bibitem{He:2016ab}
K.~He, X.~Zhang, S.~Ren, \& J.~Sun,
\newblock ``Deep residual learning for image recognition",
\newblock  {\em Proc. CVPR}, Las Vegas, 2016.

\bibitem{Srivastava:2015ab}
R.K.~Srivastava, K.~Greff, \& J.~Schmidhuber,
\newblock ``Highway networks",
\newblock  {\em arXiv.org}, 1505.00387, 2015.

\bibitem{Srivastava:2016ab}
J.G.~Zilly, R.K.~Srivastava, J.~Koutn\'{i}k, \& J.~Schmidhuber, 
\newblock ``Recurrent highway networks",
\newblock  {\em arXiv.org}, 1607.03474, 2016.

\bibitem{Zhang:2016ab}
Y.~Zhang, G.~Chen, D.~Yu, K.~Yao, S.~Khudanpur, \& J.~Glass,
\newblock ``Highway long short-term memory {RNN}s for distant speech recognition",
\newblock  {\em Proc. ICASSP}, Shanghai, 2016.

\bibitem{Tian:2016ab}
Y.~Wang \& F.~Tian,
\newblock ``Recurrent residual learning for sequence classification",
\newblock  {\em Proc. EMNLP}, Austin, 2016.

\bibitem{vandenOord:2016ab}
A.~van den Oord, N.~Kalchbrenner, \& K.~Kavukcuoglu,
\newblock ``Pixel recurrent neural networks",
\newblock  {\em Proc. ICML}, New York, 2016.


\bibitem{Pundak:2017ab}
G.~Pundak \& T.N.~Sainath,
\newblock ``Highway-{LSTM} and recurrent highway networks for speech recognition",
\newblock  {\em Proc. Interspeech}, Stockholm, 2017.

\bibitem{Kim:2017ab}
J.~Kim, M.~El-Khamy, \& J.~Lee,
\newblock ``Residual {LSTM}: {D}esign of a deep recurrent architecture for distant speech
recognition",
\newblock  {\em Proc. Interspeech}, Stockholm, 2017.

\bibitem{Baskar:2017ab}
M.K.~Baskar, M.~Karafi\'{a}t, L.~Burget, K.~Vesel\'{y}, F.~Gr\'{e}zl, \& J.H.~\v{C}ernock\'{y},
\newblock ``Residual memory networks: {F}eed-forward approach to learn long-term temporal dependencies",
\newblock  {\em Proc. ICASSP}, New Orleans, 2017.


\bibitem{Sak:2014ab}
H.~Sak, A.~Senior, \& F.~Beaufays,
\newblock ``Long short-term memory recurrent neural network architectures for large scale acoustic modeling",
\newblock  {\em Proc. Interspeech}, Singapore, 2014.

\bibitem{Bengio:1993ab}
Y.~Bengio \& P.~Frasconi,
\newblock {\em Creadit assignment through time: {A}lternatives to backpropagation},
\newblock  {\em Advances in NIPS 6}, Hong Kong, 1993.





   
\bibitem{Schuster:1997ab}
M.~Schuster \& K.K.~Paliwal,
 \newblock ``Bidirectional recurrent neural networks",
 \newblock {\em IEEE Transactions on Signal Processing},
   vol. 45, pp. 2673--2681, 1997.

 \bibitem{Young:2015ab}
 S.~Young, G.~Evermann, M.~Gales, T.~Hain, D.~Kershaw, X.~Liu, G.~Moore,
   J.~Odell, D.~Ollason, D.~Povey, A.~Ragni, V.~Valtchev, P.~Woodland, \& C.~Zhang,
 \newblock {\em The {HTK} Book (for {HTK} version 3.5)},
 \newblock Cambridge University Engineering Department, 2015.
 
 \bibitem{Zhang:2015ef}
 C.~Zhang \& P.C.~Woodland,
 \newblock ``A general artificial neural network extension for {HTK}",
 \newblock  {\em Proc. Interspeech}, Dresden, 2015.

\bibitem{Zhang:2017ab}
 C.~Zhang,
\newblock {\em Joint Training Methods for Tandem and Hybrid Speech Recognition Systems using Deep Neural Networks},
\newblock  {Ph.D. thesis}, University of Cambridge, Cambridge, UK, 2017.

\bibitem{Lin:1996ab}
T.~Lin, B.G.~Horne, P.~Ti\v{n}o, \& C. Lee Giles,
\newblock ``Learning long-term dependencies in {NARX} recurrent neural networks",
 \newblock {\em IEEE Transactions on Neural Networks},
   vol. 7, pp. 1329--1338, 1996.

\bibitem{Tino:2004ab}
P.~Ti\v{n}o, M.~\v{C}er\v{n}ansk\'{y}, \& L.~Be\v{n}u\v{s}kov\'{a},
\newblock ``Markovian architectural bias of recurrent neural networks",
 \newblock {\em IEEE Transactions on Neural Networks},
   vol. 15, pp. 6--15, 2004.
   
\bibitem{Sutskever:2010ab}
I.~Sutskever \& G.~Hinton,
\newblock ``Temporal-kernel recurrent neural networks",
 \newblock {\em Neural Networks},
   vol. 23, pp. 239--243, 2010.


\bibitem{Soltani:2016ab}
R.~Soltani \& H.~Jiang,
\newblock ``Higher order recurrent neural networks",
\newblock  {\em arXiv.org}, 1605.00064, 2016.

 \bibitem{Huang:2017ab}
 H.~Huang \& B.~Mak,
 \newblock ``To improve the robustness of {LSTM-RNN} acoustic models using higher-order feedback from multiple histories",
 \newblock  {\em Proc. Interspeech}, Stockholm, 2017.

\bibitem{mgb3website} \url{http://www.mgb-challenge.org}

 \bibitem{Bell:2015ab}
P.~Bell, M.J.F.~Gales, T.~Hain, J.~Kilgour, P.~Lanchantin, X.~Liu, A.~McParland, S.~Renals, O.~Saz, M.~Wester, \& P.C.~Woodland,
 \newblock ``The {MGB} challenge: {E}valuating multi-genre broadcast media
transcription",
 \newblock  {\em Proc. ASRU}, Scottsdale, 2015.

 \bibitem{Lanchantin:2016ab}
P.~Lanchantin, M.J.F.~Gales, P.~Karanasou, X.~Liu, Y.~Qian, L.~Wang, P.C.~Woodland, \& C.~Zhang,
 \newblock ``Selection of {M}ulti-{G}enre {B}roadcast data for the training of
automatic speech recognition systems",
 \newblock  {\em Proc. Interspeech}, San Francisco, 2016.
 

 \bibitem{Richmond:2010ab}
K.~Richmond, R.~Clark, \& S.~Fitt,
 \newblock ``On generating {C}ombilex pronunciations via morphological analysis",
 \newblock  {\em Proc. Interspeech}, Makuhari, 2010. 
 
 
 \bibitem{Evermann:2000ab}
G.~Evermann \& P.~Woodland,
 \newblock ``Large vocabulary decoding and confidence estimation using word posterior probabilities",
 \newblock  {\em Proc. ICASSP}, Istanbul, 2000.

\bibitem{Woodland:2015ab}
P.C.~Woodland, X.~Liu, Y.~Qian, C.~Zhang, M.J.F.~Gales, P.~Karanasou, P.~Lanchantin, \& L.~Wang,
\newblock ``Cambridge University transcription systems for the {M}ulti-{G}enre {B}roadcast challenge",
\newblock  {\em Proc. ASRU}, Scottsdale, 2015.

\bibitem{Li:2017ab}
B.~Li \& T.N.~Sainath,
\newblock ``Reducing the computational complexity of twodimensional {LSTM}s",
\newblock  {\em Proc. Interspeech}, Stockholm, 2017.

\bibitem{Zhang:2015cd}
C.~Zhang \& P.C.~Woodland,
\newblock ``Parameterised sigmoid and {ReLU} hidden activation functions for {DNN} acoustic modelling",
\newblock  {\em Proc. Interspeech}, Dresden, 2015.




 \bibitem{Kalchbrenner:2016ab}
 N.~Kalchbrenner, I.~Danihelka, \& A.~Graves,
 \newblock ``Grid long short-term memory",
 \newblock  {\em Proc. ICLR}, San Juan, 2016.

 \bibitem{Kreyssig:2018ab}
F.L.~Kreyssig, C.~Zhang, \& P.C.~Woodland,
 \newblock ``Improved {TDNN}s using deep kernels and frequency dependent {G}rid-{RNN}s",
 \newblock  {\em Proc. ICASSP}, Calgary, 2018.
 




\end{thebibliography}
\end{document}